%% file: ijcai23.tex
\documentclass{article}
\usepackage{mathtools}
\usepackage{times}
\usepackage{soul}
\usepackage{url}

\usepackage[utf8]{inputenc}
\usepackage[small]{caption}
\usepackage{graphicx}
\usepackage{amsmath}
\usepackage{amsthm}
\usepackage{booktabs}
\usepackage{algorithm}
\usepackage[switch]{lineno}

\usepackage[noend]{algpseudocode}
\usepackage{multirow}
\usepackage{diagbox}
\usepackage{tikz}
\usetikzlibrary{plotmarks}
\usetikzlibrary{patterns}
\usepackage{stix}

\usepackage{enumitem}

\usepackage{dsfont}
\usepackage{xcolor}

\usepackage{changes}

\usepackage[hidelinks]{hyperref}

\usepackage{ijcai23}

\newtheorem{problem}{Problem}

\newtheorem{prop}{Proposition}
\newtheorem{definition}{Definition}

\title{Optimal Decision Trees For Interpretable Clustering with Constraints (Extended Version)}

\author{
Pouya Shati $^{1,3}$ \and Eldan Cohen $^2$ \And Sheila McIlraith $^{1,3}$
\affiliations
$^1$Department of Computer Science, University of Toronto, Toronto, Canada\\
$^2$Department of Mechanical and Industrial Engineering, University of Toronto, Toronto, Canada\\
$^3$Vector Institute, Toronto, Canada
\emails
\{pouya, sheila\}@cs.toronto.edu,
ecohen@mie.utoronto.ca
}

\begin{document}

\maketitle

\begin{abstract}
    Constrained clustering is a semi-supervised task that employs a limited amount of labelled data, formulated as constraints, to incorporate domain-specific knowledge and to significantly improve clustering accuracy. Previous work has considered exact optimization formulations that can guarantee optimal clustering while satisfying all constraints, however these approaches lack interpretability. Recently, decision-trees have been used to produce inherently interpretable clustering solutions, however existing approaches do not support clustering constraints and do not provide strong theoretical guarantees on solution quality. In this work, we present a novel SAT-based framework for interpretable clustering that supports clustering constraints and that also provides strong theoretical guarantees on solution quality. We also present new insight into the trade-off between interpretability and satisfaction of such user-provided constraints. Our framework is the first approach for \textit{interpretable} and \textit{constrained} clustering. Experiments with a range of real-world and synthetic datasets demonstrate that our approach can produce high-quality and interpretable constrained clustering solutions.
\end{abstract}

\input{macros-admin.tex}

\section{Introduction}

Clustering is a core unsupervised machine learning problem that aims to partition a dataset into subgroups of similar data points. In practice, it is often used to discover meaningful sub-populations such as customer segments \cite{wu2011customer} or groups of correlated genes \cite{thalamuthu2006evaluation}. Constrained clustering is a semi-supervised learning task that exploits small amounts of supervision, provided in the form of constraints, to incorporate domain-specific knowledge and to significantly improve clustering performance \cite{wagstaff2000clustering,wagstaff2001constrained}. 
The most popular types of clustering constraints are so-called \emph{instance-level pairwise must-link} constraints, \emph{cannot-link} constraints, and additional types of constraints that can be translated to such pairwise constraints \cite{davidson2005clustering,liu2015clustering}.
In the past decades, the topic of constrained clustering has received significant attention and different constrained clustering algorithms have been proposed \cite{bilenko2004integrating,pelleg2007k,liu2017partition,cohen2020ising}. In particular, approaches that are based on exact optimization formulations, such as integer programming, constraint programming, and satisfiability have obtained state-of-the-art performance \cite{davidson2010sat,berg2017cost,DAO201770,babaki2014constrained}.



%
%
%
%
Our interest is in developing an approach to constrained clustering that yields interpretable solutions with strong solution quality guarantee.
Recent approaches to clustering via decision trees have yielded a degree of interpretability by exposing clustering rationale in the branching structure of the tree \cite{frost2020exkmc,moshkovitz2020explainable,bertsimas2021interpretable,gamlath2021nearly}.
However, these approaches do not support the incorporation of clustering constraints. Moreover, little is known about the compatibility of such clustering constraints with decision-tree clustering. \textcolor{black}{While optimal decision trees have received significant attention in recent years, most of the literature is focused on classification tasks \cite{ignatiev2021reasoning}.}


In this work, we present the first approach for \textit{interpretable} and \textit{constrained} optimal clustering based on decision trees. We make the following contributions:
\begin{enumerate}
    \item We present a novel SAT-based encoding of optimal constrained clustering that is interpretable by a decision tree. Our formulation supports two well-known clustering objectives as well as pairwise instance-level clustering constraints. To our knowledge this is the first approach for inherently-interpretable constrained clustering.
    \item We introduce novel techniques for efficient encoding of clustering problems and for improving search performance. Specifically, we introduce the notion of distance classes to support bounded suboptimal clustering, and a set of pruning rules to reduce the number of clauses.
    \item We empirically evaluate our approach over real-world and synthetic datasets and show that it leads to high-quality, inherently-interpretable clustering solutions that satisfy a given set of pairwise constraints.
    \item We present theoretical and empirical results on the trade-off between interpretability via decision trees and satisfaction of user-provided clustering constraints.
\end{enumerate}

\section{Preliminaries}


\subsection{Decision Trees for Constrained Clustering}\label{sec:tree_clustering}
\begin{definition}[Decision Tree]\label{def:dt}
Given a set of features $F$ and number of clusters $k$, a decision tree $\mathcal{D}$ is a set of branching nodes $\mathcal{T}_B$, a set of leaf nodes $\mathcal{T}_L$, a root node $\delta\in \mathcal{T}_B$, a parent function $p:\mathcal{T}_B\cup \mathcal{T}_L \rightarrow \mathcal{T}_B$, left and right child functions $l,r:\mathcal{T}_B \rightarrow \mathcal{T}_B\cup\mathcal{T}_L$, a node feature selection function $\beta: \mathcal{T}_B\rightarrow F$ together with a threshold selection function $\alpha: \mathcal{T}_B\rightarrow dom(F)$, and finally a leaf labelling function $\theta: \mathcal{T}_L \rightarrow [1..k]$.
\end{definition}

\begin{definition}[Tree Cluster Assignment]\label{def:tree_cl_assign}
Given a dataset $X$, and the number of clusters $K$, a decision tree $\mathcal{D}$ direct each point $x\in X \subset R^{|F|}$ to one of its leaves by starting from the root and recursively moving to its left or right child, depending on whether the value of the chosen feature is less than or equal to the threshold, or not. The point is then assigned the cluster label of the leaf, represented by $\Theta_\mathcal{D}(x)$.

The tree clustering $\Theta_\mathcal{D}$ partitions $X$ into $k$ clusters such that each cluster is non-empty,
$$\sum_{x_i \in X} \mathds{1}[\Theta_D(x_i) = k] \geq 1 \qquad \forall k \in 1..K.$$
Consistent with recent work \cite{frost2020exkmc}, each cluster is allowed to spread across multiple leaves to improve expressiveness and the ability to satisfy user-provided constraints.
If two points are in the same cluster (resp. different clusters), they are said to be clustered together (resp. separately).

\end{definition}

\begin{definition}[Pairwise Clustering Constraints]\label{def:cl_constraint}
Given a dataset $X$, a set of pairs of data points named must-link constraints $ML$, and a set of pairs of data points named cannot-link constraints $CL$, a clustering $\Theta_\mathcal{D}$ is said to respect the constraints if it clusters all of must-link pairs together and all of cannot-link pairs separately:
\begin{align}
    & \forall (x_1,x_2)\in ML: \Theta_\mathcal{D}(x_1) = \Theta_\mathcal{D}(x_1) \\
    & \forall (x_1,x_2)\in CL: \Theta_\mathcal{D}(x_1) \neq \Theta_\mathcal{D}(x_1)
\end{align}
\end{definition}

\begin{definition}[Minimum Split and Maximum Diameter]\label{def:ms-md}
Given a dataset $X$, the number of clusters $k$, and a clustering $\Theta_\mathcal{D}:X\rightarrow 1..k$, the minimum split is the shortest distance between two points that are clustered separately,
\begin{equation}
MS_\mathcal{D} = min(\{|x_1-x_2|\, |\, \Theta_\mathcal{D}(x_1) \neq \Theta_\mathcal{D}(x_2)\},
\end{equation}
and the maximum diameter is the longest distance between two points that are clustered together,
\begin{equation}
MD_\mathcal{D} = max(\{|x_1-x_2|\, |\, \Theta_\mathcal{D}(x_1) = \Theta_\mathcal{D}(x_2)\}.
\end{equation}
\end{definition}
We assume $|x_i-x_j|$ to be the Euclidean distance, however any totally ordered measure of distance is applicable.



\subsection{Problem Definitions}\label{sec:problem_def}
We consider two well-known clustering objectives \cite{DAO201770}: (1) minimizing the maximum diameter (MD); (2) a bi-criteria objective that consists of minimizing the maximum diameter and maximizing the minimum split ([MD,MS]). Specifically, we consider bounded approximations parameterized by $\epsilon$ such that $\epsilon=0$ indicates an optimal solution (or Pareto optimal for [MD,MS]).


\begin{problem}[MD $\epsilon$-Optimal Tree Clustering]\label{prob:md}
Given a dataset $X$, number of clusters $k$, a set of must-links constraints $ML$, a set of cannot-links constraints $CL$, decision tree depth $d$, and approximation parameter $\epsilon$, find a complete decision tree $\mathcal{D}$ of depth $d$ such that:
\begin{enumerate}
    \item $\Theta_\mathcal{D}$ respects the constraints $ML$ and $CL$;
    \item $MD_\mathcal{D} \leq MD_\mathcal{D^*} + \epsilon$,
\end{enumerate} 
where $\mathcal{D^*}$ is an optimal solution with respect to the objective of minimizing the maximum diameter.
\end{problem}


\begin{problem}[MS-MD $\epsilon$-Pareto Optimal Tree Clustering]\label{prob:ms-md}
Given a dataset $X$, number of clusters $k$, a set of must-links constraints $ML$, a set of cannot-links constraints $CL$, decision tree depth $d$, and approximation parameter $\epsilon$, find a complete decision tree $\mathcal{D}$ of depth $d$ such that:
\begin{enumerate}
    \item $\Theta_\mathcal{D}$ respects the constraints $ML$ and $CL$;
    \item $MS_\mathcal{D} \geq MS_\mathcal{D^*} - \epsilon$;
    \item $MD_\mathcal{D} \leq MD_\mathcal{D^*} + \epsilon$,
\end{enumerate}
where $\mathcal{D^*}$ is a Pareto optimal solution with respect to the bi-criteria objective of maximizing the minimum split and minimizing the maximum diameter.
\end{problem}



Note that Problem~\ref{prob:md} and Problem~\ref{prob:ms-md} are not always feasible since a decision tree of a given depth cannot represent all possible clusterings, and may not be able to represent any clustering that satisfies the $ML$ and $CL$ constraints. However in Proposition~\ref{prop:completeness} we show that for every clustering there exists a tree of certain depth that can represent it.\footnote{We only consider cases where the ML and CL constraints are consistent, i.e., where there exists a clustering that satisfies these constraints. For example, if the same pair of points is found in both the $ML$ and $CL$ constraints, there exists no clustering that would satisfy these constraints.}

\begin{prop}[Decision Tree Completeness]\label{prop:completeness}
Given a dataset $X$ and an arbitrary clustering $\Theta_C$ such that
$$\forall x_1, x_2{:} (\Theta_C(x_1) \neq \Theta_C(x_2)) \rightarrow (\exists j\in F: x_1[j] \neq x_2[j])$$
there exists a complete decision tree $\mathcal{D}$ of sufficiently high depth $d$, which partitions $X$ into the same clusters $\forall x\in X: \Theta_\mathcal{D}(x) = \Theta_C(x)$.\footnote{All proofs appear in the appendix.}
\end{prop}




\section{SAT-based $\epsilon-$Optimal Tree Clustering}\label{sec:approach}

In this section, we present our approach for solving the problems in Section~\ref{sec:problem_def}. First, we present a way to simplify our handling of pairs while still maintaining the approximation guarantee. Then, we present a novel SAT-based encoding for interpretable and constrained tree clustering.

\subsection{Distance Classes} \label{sec:distance_classes}

To encode the $\epsilon$-approximation in Problem~\ref{prob:md} and Problem~\ref{prob:ms-md}, we divide the set of all pairs of data points based on their distance into $\mu$ non-overlapping intervals called distance classes $\mathcal{D}=\{D_1, D_2, ..., D_\mu\}$ such that the smallest and largest distances in each class are less than $\epsilon$ apart. Specifically, we employ a greedy procedure that sorts all pairs from the smallest distance to the largest and greedily adds them one by one, creating new classes as needed to guarantee that the distances in each class are at most $\epsilon$ apart. All pairs in the same class are treated similarly w.r.t. being clustered together or not.

\textcolor{black}{In order to maximize the minimum split, we consider the index $\lambda^+$ such that any pair of data points in distance classes $D_1..D_{\lambda^+}$ must be clustered together (Eq. \eqref{eq:lambda_plus}). Similarly, to minimize the maximum diameter, we consider the index $\lambda^-$ such that any pair in distance classes $D_{\lambda^-{+}1}..D_\mu$ must be clustered separately (Eq. \eqref{eq:lambda_minus}).}
\begin{align}
    ((x_1,x_2)\in D_w, w \leq \lambda^+) & \rightarrow \Theta_C(x_1) = \Theta_C(x_2) \label{eq:lambda_plus} \\
    ((x_1,x_2)\in D_w, w > \lambda^-) & \rightarrow \Theta_C(x_1) \neq \Theta_C(x_2) \label{eq:lambda_minus} 
\end{align}
Note that in any feasible clustering, for all $\lambda^+$ and $\lambda^-$ values that satisfy (Eqs. \eqref{eq:lambda_plus} and \eqref{eq:lambda_minus}), we have that $MS \geq min(\{ |x_1 - x_2| \, | \, (x_1,x_2)\in D_{\lambda^++1} \})$ and $MD \leq max(\{ |x_1 - x_2| \, | \, (x_1,x_2)\in D_{\lambda^-} \})$. We therefore focus on optimizing the indices $\lambda^+$ and $\lambda^-$ to obtain $\epsilon$-optimal solutions to Problem~\ref{prob:md} and Problem~\ref{prob:ms-md} (Proposition~\ref{prop:eps_approx}).

\begin{prop}[MD and MS-MD $\epsilon$-approximation]\label{prop:eps_approx}
Let $\Theta_\mathcal{D}$ be a tree cluster assignment. We have that: 
\begin{enumerate}
    \item If $\Theta_\mathcal{D}$ is an optimal solution w.r.t minimizing $\lambda^-$ then $\Theta_\mathcal{D}$ is an $\epsilon$-optimal solution w.r.t the MD objective.
    \item If $\Theta_\mathcal{D}$ is a Pareto-optimal solution w.r.t minimizing $\lambda^-$ and maximizing $\lambda^+$ then $\Theta_\mathcal{D}$ is an $\epsilon$-Pareto optimal solution w.r.t the bi-criteria MD-MS objective.
\end{enumerate}
\end{prop}



\subsection{MaxSAT Encoding}\label{sec:encoding}

A SAT formula is a conjunction of clauses, a clause is a disjunction of literals, and a literal is either a Boolean variable or its negation. A clause is \emph{satisfied} if at least one of its literals is true. The SAT problem consists of finding an assignment of the variables that satisfies all clauses in a formula \cite{biere2009handbook}. We model the construction of $\epsilon$-optimal constrained clustering trees as Partial MaxSAT, an optimization variant of SAT that divides clauses into \emph{hard} and \emph{soft} clauses, requiring variable assignments to satisfy all hard clauses and to maximize the number of satisfied soft clauses.


\paragraph{Variables.} 
Table~\ref{tbl:variables} describes the set of Boolean variables.

\begin{table}[h]\caption{Boolean Variables in our Model}\label{tbl:variables}
\begin{tabular}{p{0.5cm}|p{7cm}}
\toprule
$a_{t,j}$ & Feature $j$ is chosen for the split at branching node~$t$ \\ \midrule
$s_{i,t}$ & {Point $i$ is directed towards the left child, if it passes through branching node $t$} \\ \midrule
$z_{i,t}$ & Point $i$ ends up at leaf node $t$ \\ \midrule
$g_{t,c}$ & The cluster assigned to leaf $t$ is or comes after $c$ \\ \midrule
$x_{i,c}$ & The cluster assigned to point $i$  is or comes after $c$ \\ \midrule
$b^-_{w}$ & {(The negation of) whether the pairs in  distance class $w$ should be clustered separately} \\ \midrule
$b^{+}_{w}$ & The pairs in class $w$ should be clustered together \\ \bottomrule
\end{tabular}
\end{table}

\paragraph{Clauses.}\label{sec:encoding_clauses}



We encode the construction of the decision tree (Eqs. \eqref{eq:1}-\eqref{eq:9}) following \citeauthor{shati2021sat} \shortcite{shati2021sat}, a state-of-the-art SAT-based decision-tree classifier that naturally supports numeric features that are prevalent in clustering problems. These clauses guarantee that exactly one feature is chosen at each branching node (Eqs.\ \eqref{eq:1}-\eqref{eq:2}), the points are directed to the left or right child of each branching node based on their value of the chosen feature (Eqs.\ \eqref{eq:3}-\eqref{eq:4}), the appearance of points at leaves correctly correspond to the path that they are directed through within the tree (Eqs.\ \eqref{eq:5}-\eqref{eq:7}), and thresholds are non-trivial (Eqs.\ \eqref{eq:8}-\eqref{eq:9}).\footnote{The set $O_j(X)$ contains all consecutive pairs of points when ordered according to feature $j$ and the set $A_l(t)$ ($A_r(t)$) contains all nodes that have $t$ as descendent of their left (right) child.}\footnote{For a detailed description of the clauses in Eqs.\ \eqref{eq:1}-\eqref{eq:9}, we refer the reader to \citeauthor{shati2021sat} \shortcite{shati2021sat}.}

{
\small\begin{align}
& (\lnot a_{t,j}, \lnot a_{t,j'}) & \forall t\in\mathcal{T}_B, j\neq j'\in F \label{eq:1}  \\ \hline
& (\bigvee_{j \in F}a_{t,j}) & \forall t\in\mathcal{T}_B \label{eq:2} \\ \hline
& (\lnot a_{t,j}, s_{i,t}, \lnot s_{i',t}) & \forall t\in\mathcal{T}_B, j\in F, (i,i')\in O_j(X) \label{eq:3} \\ \hline
\begin{split}
\mathrlap{(\lnot a_{t,j}, \lnot s_{i,t}, s_{i',t})} \\
\mathrlap{\forall t\in\mathcal{T}_B, j\in F, (i,i')\in O_j(X), x_i[j]=x_{i'}[j]} \label{eq:4}
\end{split} 
\\ \hline
& (\lnot z_{i,t}, s_{i,t'}) & \forall t\in\mathcal{T}_L, x_i\in X, t'\in A_l(t)\label{eq:5} \\ \hline
&  (\lnot z_{i,t}, \lnot s_{i,t'}) & \forall t\in\mathcal{T}_L, x_i\in X, t'\in A_r(t) \label{eq:6} \\ \hline
\begin{split}
\mathrlap{(z_{i,t}, \bigvee_{t' \in A_l(t)} \lnot s_{i, t'}, \bigvee_{t' \in A_r(t)} s_{i, t'})} \\
\mathrlap{\forall t\in\mathcal{T}_L, x_i\in X} \label{eq:7}
\end{split} \\ \hline
 & (\lnot a_{t,j}, s_{\#_j^1,t}) & \forall t\in\mathcal{T}_B, j\in F \label{eq:8} \\ \hline
 &  (\lnot a_{t,j}, \lnot s_{\#_j^{|X|},t}) & \forall t\in\mathcal{T}_B, j\in F \label{eq:9}
\end{align}
}


The following clauses extend the basic decision tree encoding to support $\epsilon$-optimal clustering trees that satisfy the $ML$ and $CL$ constraints.
The clauses in Eq. \eqref{eq:10} guarantee well-formed unary encoding of cluster labels in each leaf.\footnote{A unary encoding is well-formed if it does not include the sequence 01 at any point.}
\begin{align}
& (g_{t,c}, \lnot g_{t,c+1}) \qquad \qquad & \forall t\in\mathcal{T}_L, c\in[1..k-2] \label{eq:10}
\end{align}


Eqs. (\ref{eq:11})-(\ref{eq:12}) guarantee that the label assigned to each data point matches the label of leaf the data point reaches.
\begin{align}
& (\lnot z_{i,t} ,\lnot g_{t,c}, x_{i,c}) & \forall t\in\mathcal{T}_L, x_i\in X, c\in[1..k-1] \label{eq:11} \\
& (\lnot z_{i,t} ,g_{t,c}, \lnot x_{i,c}) & \forall t\in\mathcal{T}_L, x_i\in X, c\in[1..k-1] \label{eq:12}
\end{align}

Eqs. \eqref{eq:13}-\eqref{eq:14} break the ties between the equivalent cluster assignments. We consider two cluster assignments $\Theta^1$ and $\Theta^2$ equivalent if there exists a relabelling, $\Gamma: K \rightarrow K$, such that $\Gamma \circ \Theta^1 = \Theta^2$. To break ties, we force each $x_i$ in the ascending order to be assigned to the first empty cluster, if it needs a new one. This guarantees that there are no two feasible solutions that are relabellings of each other. Note that we do not eliminate viable solutions, since any clustering can be renamed into an equivalent one that respects this property. 
\begin{align}
& (\lnot x_{c,c}) & \forall c\in[1..k-1] \label{eq:13} \\
& (\lnot x_{i,c}, \bigvee_{i' < i} x_{i', c-1}) & \forall x_i\in X, c\in[2..k-1], c<i \label{eq:14}
\end{align}

The clause in Eq.\ (\ref{eq:15}) alongside the tie-breaking ones guarantee that all clusters are non-empty, assuming $|X| \geq K$. If we substitute $k$ with $k'<k$ in Eq.\ (\ref{eq:15}), $k'$ clusters are guaranteed to be non-empty. Thus, we can enforce minimum $k'$ and maximum $k$ clusters. In our experiments we focus on the setting where all clusters are non-empty.
\begin{align}
& (\bigvee_{i} x_{i, k-1}) & \label{eq:15}
\end{align}

Eqs.\ \eqref{eq:16}-\eqref{eq:18}, namely the unconditional separating clauses, guarantee that pairs in $CL$ are clustered separately.
\begin{align}
& (x_{i,1}, x_{i',1}) & \forall (i,i')\in CL \label{eq:16} \\
\hline & (\lnot x_{i,k-1}, \lnot x_{i',k-1}) & \forall (i,i')\in CL \label{eq:17} \\
 \hline \begin{split}
& (\lnot x_{i,c}, \lnot x_{i',c}, x_{i,c+1}, x_{i',c+1}) \qquad \\ 
& \forall (i,i')\in CL, c\in[1..k-2] 
\end{split} \label{eq:18}
\end{align}

Eqs.\ \eqref{eq:19}-\eqref{eq:20}, namely the unconditional co-clustering clauses, guarantee that pairs in $ML$ are clustered together.
\begin{align}
& (\lnot x_{i,c}, x_{i',c}) \qquad & \forall (i,i')\in ML, c\in[1..k-1] \label{eq:19} \\
& (x_{i,c}, \lnot x_{i',c}) & \forall (i,i')\in ML, c\in[1..k-1] \label{eq:20}
\end{align}

Eqs.\ \eqref{eq:21}-\eqref{eq:23}, namely the conditional separating clauses, guarantee that a $b^-_w$ variable being set to false forces the pairs in distance class $w$ to be clustered separately.
\begin{align}
& (b^-_w, x_{i,1}, x_{i',1}) & \forall D_w \in \mathcal{D}, (i,i')\in D_w \label{eq:21} \\ \hline
& (b^-_w, \lnot x_{i,k-1}, \lnot x_{i',k-1}) & \forall D_w \in \mathcal{D}, (i,i')\in D_w \label{eq:22} \\ \hline
\begin{split}
\mathrlap{(b^-_w, \lnot x_{i,c}, \lnot x_{i',c}, x_{i,c+1}, x_{i',c+1})} \\
\mathrlap{\forall D_w \in \mathcal{D}, (i,i')\in D_w, c\in[1..k-2]}  
\end{split} \label{eq:23}
\end{align}

Eqs.\ \eqref{eq:24}-\eqref{eq:25}, namely the conditional co-clustering clauses, guarantee that a $b^+_w$ variable being set to true forces the pairs in distance class $w$ to be clustered together.
\begin{align}
\begin{split}
& (\lnot b^+_w, \lnot x_{i,c}, x_{i',c}) \qquad\qquad \\
& \forall D_w \in \mathcal{D}, (i,i')\in D_w, c\in[1..k-1] \qquad\qquad 
\end{split} \label{eq:24} \\
\hline \begin{split}
& (\lnot b^+_w, x_{i,c}, \lnot x_{i',c}) \\
& \forall D_w \in \mathcal{D}, (i,i')\in D_w, c\in[1..k-1] \label{eq:25}
\end{split}
\end{align}

Eqs.\ \eqref{eq:26}-\eqref{eq:28} guarantee that the distance classes are partitioned according to valid $\lambda^+$ and $\lambda^-$ values, i.e.,  that $D_1..D_{\lambda^+}$ distance classes are clustered together, and $D_{\lambda^-{+}1}..D_\mu$ distance classes are clustered separately.
\begin{align}
& (\lnot b^-_w, b^-_{w-1}) & \qquad\qquad\qquad & \forall D_w \in \mathcal{D}, w>1 \label{eq:26} \\
& (\lnot b^+_w, b^+_{w-1}) & \qquad\qquad\qquad & \forall D_w \in \mathcal{D}, w>1 \label{eq:27} \\
& (\lnot b^+_w, b^-_w) & \qquad\qquad\qquad & \forall D_w \in \mathcal{D} \label{eq:28}
\end{align}

\paragraph{Smart Pairs.}\label{sec:smart_pairs}
Since each pair of points has corresponding clauses that handle being clustered together or separately, a naive encoding will be quadratic in the number of points $|X|$. This is significant since all the other parts of the encoding are linear in $|X|$. To reduce the number of clauses, we see the set of points as nodes in a graph and exploit connections between pairs. Pairs that are forced to be clustered together are represented by \textit{positive} edges and pairs that are force to be clustered separately by \textit{negative} edges. The positive edges imply a set of connected components of points that will be clustered together while a negative edge between nodes in different components indicates that each of the components are mutually exclusive, i.e., each component will be in a different cluster. The order $<^*$, which sorts the pairs based on distance and breaks ties arbitrary, is used to greedily build the set of positive edges ($E^+$) and the set of negative edges ($E^-$). As we incrementally build the sets, each new edge can be classified as \textit{inner} or \textit{crossing}, based on the previous members in $E^+$ and $E^-$, to help us detect infeasibility or redundancy of clauses.


\begin{definition}[Inner and Crossing edges]\label{def:inner_cross_edge}
Given the sets $E^+$ and $E^-$, a new edge is said to be an:
\begin{itemize}
    \item Inner edge, if it connects two nodes within an existing connected component based on $E^+$.
    \item Crossing edge, if it connects two nodes in two connected components based on $E^+$ that are
    mutually exclusive based on $E^-$.
\end{itemize}
We will use edges and the pairs of points that they represent interchangeably onward.
\end{definition}

Since $E^+$ ($E^-$) represents the set of pairs that are forced to be clustered together (separately), an inner pair is forced to have the same label and a crossing pair to have different labels. We make use of this fact to avoid adding co-clustering clauses Eqs. (\ref{eq:19},~\ref{eq:20},~\ref{eq:24},~\ref{eq:25}) or separating clauses Eqs. (\ref{eq:16},~\ref{eq:17},~\ref{eq:18},~\ref{eq:21},~\ref{eq:22},~\ref{eq:23}) when it is redundant to do so. Furthermore, we can conclude infeasibility when an inner (crossing) pair is forced to be clustered separately (together).


Our treatment of the unconditional clauses Eqs. (\ref{eq:16}-\ref{eq:20}) and of the conditional ones Eqs. (\ref{eq:21}-\ref{eq:25}) differ in two ways. 
\begin{enumerate}
    \item For the unconditional clauses the $E^+$ and $E^-$ sets are respectively constructed from $ML$ and $CL$ links. For conditional ones however, we also include the pairs that are implied to be co-clustered (separated), due to the order of distance imposed by $\lambda^+$ and $\lambda^-$ values. Note that implied pairs for the processing of conditional co-clustering clauses are not valid for the processing of conditional separating clauses.
    \item If the unconditional separation or co-clustering of a pair is infeasible, the problem is infeasible. However, for conditional clauses, we only fix the corresponding $b^+_w$ or $b^-_w$ variable to satisfy the conditional clause.
    
\end{enumerate}

The detailed procedure is described in the appendix. 



\paragraph{Objective.}
In Section~\ref{sec:distance_classes} we established that in order to find an $\epsilon$-approximation of an MS-MD Pareto optimal solution, we need to find a solution that is Pareto optimal with regards to maximizing $\lambda^+$ and minimizing $\lambda^-$. Similarly, for an $\epsilon$-approximation of an MD optimal solution we need to find a solution that minimizes $\lambda^-$. Note that $\lambda^+$ and $\lambda^-$ are the number of $b^+_w$ and $b^-_w$ variables set to true, respectively.
\begin{align}
& \lambda^+ = \sum_w \mathds{1}(b^+_w = true)  \\
& \lambda^- = \sum_w \mathds{1}(b^-_w = true)
\end{align}

To minimize $\lambda^-$, we simply introduce a soft clause with unit weight for each $\lnot b^-_w$ (Eq. \eqref{eq:29}). To obtain a Pareto-optimal solution, we can optimize any function that is increasing w.r.t $\lambda^-$ and decreasing w.r.t $\lambda^+$ as objective. We opt for minimizing the simple combination of $\lambda^- - \lambda^+$ and model it using the soft clauses in Eq. \eqref{eq:29} and Eq. \eqref{eq:30}.
\begin{align}
& (\lnot b^-_w) & w\in W \label{eq:29} \\
& (b^+_w) & w\in W \label{eq:30}
\end{align}
For the MD objective, the $b^+_w$ variables and the clauses in Eqs. \eqref{eq:24}-\eqref{eq:25} become irrelevant and can be removed.




\section{Experiments}

\subsection{Experiment Setup}

We use the Loandra solver \cite{berg2019core} to solve our tree clustering encoding. \textcolor{black}{Loandra is an any-time solver that guarantees optimality if run to completion, but can also produce intermediate solutions.} We run experiments on a server with two 12-core Intel E5-2697v2 CPUs and 128G of RAM.

\paragraph{Datasets.}
We run experiments on seven real datasets from the UCI repository \cite{Dua:2019} and four synthetic datasets from FCPS \cite{ultsch2020fundamental}. 
The datasets vary in size, number of features, and number of clusters, as presented in Table~\ref{tab:tree}. 
For all datasets, we normalize the values of each feature in the range $[0,100]$ so that features with larger values do not dominate the pairwise distances. 


\paragraph{Constraint Generation.} We evaluate the performance of our approach for different number of constraints, relative to the dataset size. Specifically, for a given $\kappa$ value with $0\leq\kappa\leq \frac{|X|-1}{2}$, we generate a set of $\kappa\cdot|X|$ random clustering constraints following the methodology in \citeauthor{wagstaff2000clustering} \shortcite{wagstaff2000clustering}: (1) We generate $\kappa\cdot|X|$ pairs of data points without repetition; (2) For each pair, we generate a ML constraint if both data point share the same ground-truth label and a CL constraint otherwise.

\paragraph{Evaluation.} 
We evaluate the quality of the obtained clusterings based on ground-truth labels using two well-known external clustering evaluation metrics: the Adjusted Rand Index (ARI) \cite{hubert1985comparing} and the Normalized Mutual Information (NMI) \cite{strehl2002cluster}.




\begin{table*}[htp!]
    \centering
    \caption{Interpretable constrained clustering with $\epsilon=0.1$ averaged over $20$ runs.}
    \label{tab:tree}
    \small
    \begin{tabular}{cc|cccc|cccc}
    \toprule
\multirow{2}{*}{Dataset} & \multirow{2}{*}{$\kappa$} & \multicolumn{4}{c}{[MD,MS]} & \multicolumn{4}{c}{MD}\\
 & & ARI & ARI (CC) & Feas. & Time (s) & ARI & ARI (CC) & Feas. & Time (s) \\
 \midrule
Iris  & 0.0 & 0.6 & 0.6 & 20 & 0.6 & 0.62 & 0.7 & 20 & 0.4 \\
$\vert X \vert=150$ & 0.1 & 0.83 & 0.78 & 20 & 0.7 & 0.71 & 0.55 & 20 & 0.4 \\
$\vert F \vert=4$ & 0.25 & 0.86 & 0.8 & 20 & 0.7 & 0.81 & 0.6 & 20 & 0.4 \\
$K=3$ & 0.5 & 0.91 & 0.8 & 20 & 0.8 & 0.88 & 0.63 & 20 & 0.5 \\
$d=3$ & 1.0 & 0.95 & 0.91 & 16 & 0.7 & 0.94 & 0.71 & 16 & 0.4 \\
\hline Wine  & 0.0 & 0 & 0 & 20 & 0.7 & 0.38 & 0.21 & 20 & 0.6 \\
$\vert X \vert=178$ & 0.1 & 0.69 & 0.53 & 20 & 1.2 & 0.41 & 0.19 & 20 & 0.6 \\
$\vert F \vert=13$ & 0.25 & 0.79 & 0.57 & 20 & 1.8 & 0.6 & 0.2 & 20 & 0.9 \\
$K=3$ & 0.5 & 0.82 & 0.5 & 20 & 6.9 & 0.72 & 0.2 & 20 & 5.4 \\
$d=3$ & 1.0 & 0.93 & 0.72 & 20 & 8.8 & 0.89 & 0.23 & 20 & 8 \\
\hline Glass  & 0.0 & 0.22 & 0.22 & 20 & 2.1 & 0.18 & 0.26 & 20 & 1.2 \\
$\vert X \vert=214$ & 0.1 & 0.19 & 0.1 & 20 & 9.3 & 0.16 & 0.11 & 20 & 3.3 \\
$\vert F \vert=9$ & 0.25 & 0.24 & 0.06 & 20 & 61.1 & 0.16 & 0.05 & 20 & 35.1 \\
$K=7$ & 0.5 & 0.26 & 0.07 & 19 & 954.1 & 0.24 & 0.01 & 20 & 624.9 \\
$d=4$ & 1.0 & - & 0.1 & 0 & 193.8 & - & 0.01 & 0 & 200.1 \\
\hline Ionosphere  & 0.0 & 0.01 & 0.01 & 20 & 2.4 & 0.16 & 0.08 & 20 & 2 \\
$\vert X \vert=351$ & 0.1 & 0.28 & 0.11 & 20 & 33.7 & 0.15 & 0.09 & 20 & 10.2 \\
$\vert F \vert=34$ & 0.25 & 0.5 & 0.22 & 11 & 598.9 & 0.48 & 0.12 & 11 & 880.9 \\
$K=2$ & 0.5 & - & 0.38 & 0 & 29.5 & - & 0.18 & 0 & 31.4 \\
$d=3$ & 1.0 & - & 0.78 & 0 & 5.8 & - & 0.76 & 0 & 5.7 \\
\hline Seeds  & 0.0 & 0.68 & 0.66 & 20 & 1.8 & 0.57 & 0.68 & 20 & 0.5 \\
$\vert X \vert=210$ & 0.1 & 0.68 & 0.64 & 20 & 1 & 0.67 & 0.54 & 20 & 0.5 \\
$\vert F \vert=7$ & 0.25 & 0.73 & 0.63 & 20 & 1.3 & 0.72 & 0.52 & 20 & 0.7 \\
$K=3$ & 0.5 & 0.78 & 0.64 & 14 & 1.5 & 0.78 & 0.52 & 14 & 1.2 \\
$d=3$ & 1.0 & 0.9 & 0.79 & 1 & 0.7 & 0.93 & 0.72 & 1 & 0.7 \\
\hline Libras  & 0.0 & 0.21 & 0.22 & 20 & 1802.8 & 0.2 & 0.16 & 20 & 1802.9 \\
$\vert X \vert=360$ & 0.1 & 0.17 & 0.17 & 20 & 1159.1 & 0.15 & 0.12 & 20 & 941.6 \\
$\vert F \vert=90$ & 0.25 & 0.17 & 0.14 & 20 & 1080 & 0.14 & 0.1 & 20 & 681.9 \\
$K=15$ & 0.5 & 0.18 & 0.11 & 20 & 866.2 & 0.14 & 0.07 & 20 & 305.1 \\
$d=5$ & 1.0 & 0.18 & 0.11 & 20 & 1518.3 & 0.14 & 0.06 & 20 & 859.4 \\
\hline Spam  & 0.0 & 0 & 0 & 20 & 38 & -0.02 & -0.01 & 20 & 41.5 \\
$\vert X \vert=4601$ & 0.1 & - & 0.04 & 0 & 358.9 & - & 0.05 & 0 & 379.4 \\
$\vert F \vert=57$ & 0.25 & - & 0.07 & 0 & 286.1 & - & 0.13 & 0 & 312.6 \\
$K=2$ & 0.5 & - & 0.08 & 0 & 151.1 & - & 0.25 & 0 & 152.9 \\
$d=3$ & 1.0 & - & 0.51 & 0 & 77.1 & - & 0.51 & 0 & 77.9 \\
\hline Lsun  & 0.0 & 0.44 & 0.39 & 20 & 1.2 & 0.39 & 0.39 & 20 & 0.8 \\
$\vert X \vert=400$ & 0.1 & 0.95 & 0.65 & 20 & 1 & 0.74 & 0.25 & 20 & 0.7 \\
$\vert F \vert=2$ & 0.25 & 1 & 0.94 & 20 & 1 & 0.89 & 0.25 & 20 & 0.6 \\
$K=3$ & 0.5 & 1 & 1 & 20 & 0.9 & 0.96 & 0.26 & 20 & 0.6 \\
$d=3$ & 1.0 & 1 & 1 & 20 & 0.9 & 0.98 & 0.48 & 20 & 0.6 \\
\hline Chainlink & 0 & 0.12 & 1 & 20 & 3 & 0.11 & 0.06 & 20 & 2 \\
$\vert X \vert=1000$ & 0.1 & 0.89 & 1 & 17 & 5 & 0.84 & 0.01 & 17 & 5.5 \\
$\vert F \vert=3$ & 0.25 & 0.89 & 1 & 1 & 1.8 & 0.91 & 0.03 & 1 & 1.7 \\
$K=2$ & 0.5 & - & 1 & 0 & 1.3 & - & 0.05 & 0 & 1.2 \\
$d=3$ & 1.0 & - & 1 & 0 & 1 & - & 0.56 & 0 & 0.9 \\
\hline Target  & 0.0 & 0.36 & 0.36 & 20 & 5.7 & 0.33 & 0.33 & 20 & 3.7 \\
$\vert X \vert=770$ & 0.1 & 1 & 1 & 20 & 2.8 & 0.64 & 0.57 & 20 & 18.8 \\
$\vert F \vert=2$ & 0.25 & 1 & 1 & 20 & 2.7 & 0.87 & 0.45 & 20 & 9.9 \\
$K=6$ & 0.5 & 1 & 1 & 20 & 2.4 & 0.95 & 0.25 & 20 & 4.4 \\
$d=4$ & 1.0 & 1 & 1 & 20 & 2.2 & 0.99 & 0.45 & 20 & 1.9 \\
\hline WingNut  & 0.0 & 1 & 1 & 20 & 2.5 & 1 & 0.15 & 20 & 2 \\
$\vert X \vert=1016$ & 0.1 & 1 & 1 & 20 & 1.7 & 0.99 & 0.2 & 20 & 1.3 \\
$\vert F \vert=2$ & 0.25 & 1 & 1 & 20 & 1.6 & 0.99 & 0.28 & 20 & 1.2 \\
$K=2$ & 0.5 & 1 & 1 & 20 & 1.6 & 1 & 0.49 & 20 & 1.2 \\
$d=3$ & 1.0 & 1 & 1 & 20 & 1.6 & 1 & 0.82 & 20 & 1.2 \\
\bottomrule
\end{tabular}
\end{table*}

\subsection{Baselines}
\paragraph{Constrained Clustering (CC).} We compare our approach 
with optimal constrained clustering formulation that is not restricted to conform to a decision tree. To do so, we remove the tree-related components of the encoding, namely the variables $[a_{t,j}]$, $[s_{i,t}]$, $[z_{i,t}]$, $[g_{t,c}]$ and the clauses in Eqs. \eqref{eq:1}-\eqref{eq:12}. Instead, we introduce the clauses in Eq. (\ref{eq:10_b}) that guarantee a well-formed unary encoding of labels.\footnote{Note that these clauses are redundant in the encoding of tree clustering as the condition is already enforced for the leaves.}
\begin{align}
& (x_{i,c}, \lnot x_{i,c+1}) & \forall x_i\in X, c\in[1..k-2] \label{eq:10_b}
\end{align}
When used with the MD objective and $\epsilon=0.0$, this baseline is equivalent to the maximum diameter constrained clustering formulations in previous works \cite{dao2016framework,DAO201770}, i.e., it has the same set of (feasible and) optimal solutions. However, this baseline also supports $\epsilon$-approximation and the [MD,MS] objective for a fair comparison with our approach.

\paragraph{Mixed Integer Optimization (MIO).}
The most closely related work to ours is \citeauthor{bertsimas2021interpretable} \shortcite{bertsimas2021interpretable} that presents a MIO model for (unconstrained) clustering trees. While their model does not support clustering constraints, we can extend it to incorporate such constraints. 
However, \citeauthor{bertsimas2021interpretable} note that their model does not scale well and therefore do not present any experimental results and instead focus on a heuristic procedure that does not have any solution quality guarantees and cannot naturally support clustering constraints.
Due to the relevance of the MIO model to our work, we have implemented the model using the Gurobi v10 solver and extended it to support clustering constraints (detailed description of the extended MIO model can be found in the appendix). To our knowledge, this is the only approach in the literature for constructing tree-based clustering using discrete optimization.

\subsection{Results}

For our first set of experimental results presented in Table~\ref{tab:tree}, we solve the tree clustering problem for our datasets with different values of $\kappa$. We fix the value of approximation at $\epsilon=0.1$. Consistent with previous work \cite{dao2016framework,babaki2014constrained}, we set the solver time limit to 30 minutes. To avoid bias in results due to a specific set of constraints, we generate 20 random sets of constraints and report average values for the evaluation metrics and the runtime. Note that it might not be possible for a tree clustering with a fixed depth to satisfy all of the constraints, in particular in high-dimensional datasets with complex patterns where a large number of univariate splits may be needed to fit the ground-truth constraints.
Thus, we also report the number of feasible runs (out of the 20 random constraint sets) while excluding infeasible and unknown\footnote{Instances where the solver timed out without finding a feasible solution are considered ``unknown''.} cases from the computed average ARI. 

The results show that our approach can produce high quality interpretable solutions with non-zero $\epsilon$ values in the time limit. We observe that the [MD,MS] Pareto optimality objective leads to higher quality solutions compared to MD in all but a few cases without significant overhead in runtime. 


Interestingly, we observe that tree clustering almost always leads to higher quality solutions compared to CC. This seems unintuitive since CC is strictly more expressive than tree clustering. We conjecture that both clustering objectives tend to perform better with tree clustering because of its inherently restricted solution space, \textcolor{black}{resulting in potentially worse objective values but higher ARI scores.}
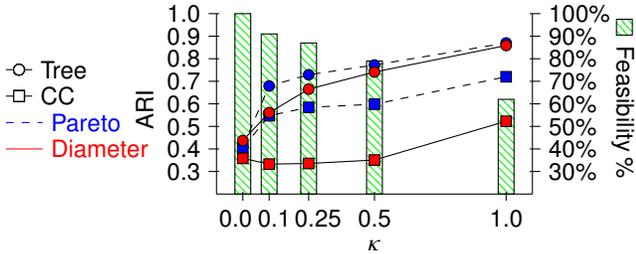
\begin{figure}[ht!]
    \caption{Results for $\epsilon=0.1$ averaged over $20$ runs and $11$ datasets.}
    \label{fig:tree}
    \begin{tikzpicture}[y=3cm, x=3.5cm,font=\sffamily]
\footnotesize
	\draw (-0.1,0.2) -- coordinate (x axis mid) (1.1,0.2);
    	\draw (-0.1,0.2) -- coordinate (y axis mid) (-0.1,1.0);
    	\draw (1.1,0.2) -- coordinate (y axis mid) (1.1,1.0);
    	\draw (0,0.21) -- (0,0.17)
			node[anchor=north] {0.0 \,\,};
		\draw (0.1,0.21) -- (0.1,0.17)
			node[anchor=north] {\, 0.1};
		\draw (0.25,0.21) -- (0.25,0.17)
			node[anchor=north] {\,\,\, 0.25};	
    	\foreach \x in {0.5,1.0}
     		\draw (\x,0.21) -- (\x,0.17)
			node[anchor=north] {\x};
    	\foreach \y in {0.3,0.4,0.5,0.6,0.7,0.8,0.9,1.0}
     		\draw (-0.09,\y) -- (-0.13,\y) 
     			node[anchor=east] {\y}; 
     			
     	\foreach \y in {30,40,50,60,70,80,90,100}
     		\draw (1.09,\y/100) -- (1.13,\y/100) 
     			node[anchor=west] {\y \%};

 	\node[below=0.5cm] at (x axis mid) {$\kappa$};
 	\node[rotate=90, left=0.95cm] at (-0.1,0.7) {ARI};
 	\node[rotate=270, right=1.2cm] at (1.1,1.0) {\quad Feasibility \%};
    \node[draw=black, pattern=north west lines, pattern color=green, right=1.11cm] at (1.1, 0.94){};
    
	\draw [draw=black, pattern=north west lines, pattern color=green]
      (-0.03,0.2) -- (0.03,0.2) -- (0.03,1.00) -- (-0.03,1.00) -- cycle;
    \draw [draw=black, pattern=north west lines, pattern color=green]
      (0.07,0.2) -- (0.13,0.2) -- (0.13,0.91) -- (0.07,0.91) -- cycle;   
    \draw [draw=black, pattern=north west lines, pattern color=green]
      (0.22,0.2) -- (0.28,0.2) -- (0.28,0.87) -- (0.22,0.87) -- cycle;    
    \draw [draw=black, pattern=north west lines, pattern color=green]
      (0.47,0.2) -- (0.53,0.2) -- (0.53,0.79) -- (0.47,0.79) -- cycle; 
    \draw [draw=black, pattern=north west lines, pattern color=green]
      (0.97,0.2) -- (1.03,0.2) -- (1.03,0.62) -- (0.97,0.62) -- cycle;  

	
	\draw [dashed] (0.0,0.410909091) -- (0.1,0.679) -- (0.25,0.729) -- (0.5,0.772222222) -- (1.0,0.87);
	
	\draw [dashed] (0,0.405454545) -- (0.1,0.547272727) -- (0.25,0.584545455) -- (0.5,0.598181818) -- (1,0.72);

	\draw plot[mark=*, mark options={fill=blue},only marks]
		coordinates {(0.0,0.410909091) (0.1,0.679) (0.25,0.729) (0.5,0.772222222) (1.0,0.87)};
	\draw plot[mark=square*, mark options={fill=blue},only marks] 
		coordinates {(0,0.405454545) (0.1,0.547272727) (0.25,0.584545455) (0.5,0.598181818) (1,0.72)};
	\draw plot[mark=*, mark options={fill=red}]
		coordinates {(0,0.437272727) (0.1,0.561) (0.25,0.665) (0.5,0.741111111) (1,0.85875)};
	\draw plot[mark=square*, mark options={fill=red}]
		coordinates {(0,0.358181818) (0.1,0.332727273) (0.25,0.335454545) (0.5,0.350909091) (1,0.522727273)}; 
 
    
	\begin{scope}[shift={(-0.9,0.4)}] 
	\draw [color=red](0,0) -- (0.14,0)
		node[right]{Diameter};
	\draw[yshift=\baselineskip,color=blue,dashed] (0,0) -- (0.14,0)
		node[right]{Pareto};
	\draw[yshift=2\baselineskip] (0,0) -- 
		plot[mark=square*, mark options={fill=white}] (0.05,0) -- (0.1,0)
		node[right]{CC};
	\draw[yshift=3\baselineskip] (0,0) -- 
		plot[mark=*, mark options={fill=white}] (0.05,0) -- (0.1,0)
    	node[right]{Tree};
	\end{scope}

\end{tikzpicture}
\end{figure}


Figure~\ref{fig:tree} 
highlights
the benefit of both tree clustering ($\circ$) and the Pareto objective (dashed). While the difference in ARI in unconstrained clustering ($\kappa=0$) is negligible, as we increase the number of constraints we observe that tree-based clustering significantly outperforms CC and the Pareto objective tends to perform better than the MD objective 
when constraints are limited. Figure~\ref{fig:tree} also shows the percentage of feasible instances vs.\ $\kappa$ (bars) and demonstrates that as problems become more constrained, it may be infeasible to find a tree clustering of a given depth that satisfies all the constraints, exposing the trade-off between interpretability via decision trees and satisfaction of user-provided clustering constraints.



Next, we investigate how depth impacts the feasibility of finding a decision tree that satisfies the clustering constraints. The results in Table~\ref{tab:infeas} show that deeper trees can resolve the infeasibility problem in some datasets, e.g., Glass and Ionosphere, however in Spam we find that even with a depth of four we are not able to find decision trees for any of the random constraint sets. Another interesting observation is that unnecessarily deep trees could lead to a lower score, demonstrating the potential benefit of restricted solution spaces that are induced by shallow trees. We observed similar trends for the NMI metric and provide the detailed results in the appendix.


\begin{table}[t!]
    \centering
    \caption{Tree clustering with the [MD,MS] objective for different tree depths ($\kappa=0.5$, $\epsilon=0.1$) averaged over $20$ runs.}
    \label{tab:infeas}
    \small
    \begin{tabular}{cc|ccc}
    \toprule
Dataset & $d$ & ARI & Feas. & Time (s) \\
 \midrule
 \multirow{3}{*}{Iris} & 2 & 0.83 & 2 & 0.3 \\
 & 3 & 0.91 & 20 & 0.8 \\
 & 4 & 0.88 & 20 & 0.8 \\
\hline \multirow{3}{*}{Wine} & 2 & 0.87 & 2 & 0.4 \\
 & 3 & 0.82 & 20 & 6.9 \\
 & 4 & 0.71 & 20 & 6.2 \\
\hline \multirow{3}{*}{Glass} & 3 & - & 0 & 2.6 \\
 & 4 & 0.26 & 19 & 953.6 \\
 & 5 & 0.22 & 20 & 35.8 \\
\hline \multirow{3}{*}{Ionosphere} & 2 & - & 0 & 0.8 \\
 & 3 & - & 0 & 29.5 \\
 & 4 & 0.69 & 18 & 731.7 \\
\hline \multirow{3}{*}{Seeds} & 2 & - & 0 & 0.3 \\
 & 3 & 0.78 & 14 & 1.5 \\
 & 4 & 0.74 & 20 & 2.1 \\
\hline \multirow{3}{*}{Libras} & 4 & 0.16 & 2 & 1801.4 \\
 & 5 & 0.18 & 20 & 866.2 \\
 & 6 & 0.17 & 20 & 315.4 \\
\hline \multirow{3}{*}{Spam} & 2 & - & 0 & 35 \\
 & 3 & - & 0 & 151.1 \\
 & 4 & - & 0 & 975.6 \\
\hline \multirow{3}{*}{Lsun} & 2 & 1 & 20 & 0.9 \\
 & 3 & 1 & 20 & 0.9 \\
 & 4 & 1 & 20 & 1.1 \\
\hline \multirow{3}{*}{Chainlink} & 2 & - & 0 & 0.8 \\
 & 3 & - & 0 & 1.3 \\
 & 4 & 1 & 20 & 3.4 \\
\hline \multirow{3}{*}{Target} & 3 & - & 0 & 1.8 \\
 & 4 & 1 & 20 & 2.5 \\
 & 5 & 1 & 20 & 4 \\
\hline \multirow{3}{*}{WingNut} & 2 & 1 & 20 & 1.5 \\
 & 3 & 1 & 20 & 1.7 \\
 & 4 & 1 & 20 & 2.1 \\
\bottomrule
\end{tabular}
\end{table}

\begin{table}[t!]\centering
    \centering
    \caption{Ablation study ([MD,MS]) averaged over $20$ runs.}
    \label{tab:ablation}
    \small
    \setlength{\tabcolsep}{5.75pt}
    \begin{tabular}{cc|rrr}
    \toprule
 Dataset & Setting & ARI & Time (s) & \# Clauses \\
 \midrule
\multirow{3}{*}{Libras} & SP \& $\epsilon{=}0.1$ & 0.18 & 866.4 & 2,082,261.2 \\
 & $\epsilon{=}0.1$ & 0.16 & 822.0 & 3,888,452.0 \\
 &  $\epsilon{=}0.0$ & 0.16 & 1197.1 & 4,140,872.0 \\
 \midrule
\multirow{3}{*}{Spam} & SP \& $\epsilon{=}0.1$ & Inf. & 151.6 & 3,823,479.2 \\
 &  $\epsilon{=}0.1$ & Inf. & 332.7 & 24,980,546.4 \\
 &  $\epsilon{=}0.0$ & Inf./Unk. & 864.0 & 69,166,751.4 \\
 \midrule
\multirow{3}{*}{WingN} & SP \& $\epsilon{=}0.1$ & 1.00 & 1.7 & 95,879.25 \\
 &  $\epsilon{=}0.1$ & 1.00 & 4.2 & 1,128,700.4 \\
 &  $\epsilon{=}0.0$ & OOM\textsuperscript{$\dagger$} & 98.3 & 3,449,740.4 \\
\bottomrule
\multicolumn{5}{l}{\textsuperscript{$\dagger$}\footnotesize{OOM indicates an out-of-memory error.}}
\end{tabular}
\end{table}

\paragraph{Comparison with MIO.}
Our experiments with the MIO baseline found that it is unable to find a feasible solution for any of the datasets for the depths specified in Table~\ref{tab:tree} across multiple runs, both in constrained and unconstrained settings. This result is consistent with \citeauthor{bertsimas2021interpretable}'s \shortcite{bertsimas2021interpretable} observation on the limited scalability of their MIO formulation and emphasizes the strong performance of our approach.

\paragraph{Ablation Study.}
Finally, we study the impact of the smart pairs procedure (Section~\ref{sec:encoding}) and the $\epsilon$-approximation (Section~\ref{sec:distance_classes}) on the performance of our approach.
Unlike the $\epsilon$-approximation, smart pairs is guaranteed to not change the set of feasible solutions and the set of optimal solutions. However, since there could be multiple optimal solutions, the different encoding may still lead to an optimal solution with a slightly different ARI score. The results presented in Table~\ref{tab:ablation} show that the aforementioned methods can reduce the number of clauses and runtimes without meaningful decrease in score.




\section{Conclusion}


In this work, we present the first approach for \textit{interpretable} and \textit{constrained} clustering using decision trees. Specifically, we present a novel SAT-based encoding for constructing clustering trees that approximate two well-known clustering objectives. Our experiments on a range of real-world and synthetic datasets demonstrate the ability of our approach to produce high-quality and interpretable clustering solutions that incorporate user-provided clustering constraints.

Our work raises several interesting questions to investigate in future work. As there are potentially many Pareto-optimal solutions, investigating and empirically evaluating 
strategies \textcolor{black}{or tools, e.g. multi-objective solvers \cite{jabs2022maxsat},} for exploring the Pareto front and selecting promising solutions is an interesting direction for future work. One of the challenges identified in this work is that highly-constrained problems in complex, high-dimensional datasets can become infeasible for a given tree depth. In future work, we would like to investigate strategies to overcome this challenge, e.g., by converting the hard clustering constraints into soft constraints that are encouraged rather than required to be satisfied.

\section*{Acknowledgments}

We gratefully acknowledge funding from the Natural Sciences and Engineering Research Council of Canada (NSERC) and the Canada CIFAR AI Chairs Program. Resources used in preparing this research were provided, in part, by the Province of Ontario, the Government of Canada through CIFAR, and companies sponsoring the Vector Institute for Artificial Intelligence (\url{www.vectorinstitute.ai/partners}). The final author would also like to thank the Schwartz Reisman Institute for Technology and Society for providing a rich multi-disciplinary research environment.

\bibliographystyle{named}
\bibliography{cluster-tree}

\appendix
\include{appendix}

\end{document}

%% file: macros-admin.tex

\newcommand{\revisit}[1]{\textcolor{blue}{#1}}
\newcommand{\revise}[1]{\textcolor{blue}{#1}}
\newcommand{\review}[1]{\textcolor{blue}{#1}}
\newcommand{\alt}[1]{\textcolor{brown}{#1}}
\newcommand{\althide}[1]{}
\newcommand{\addedsm}[1]{\textcolor{red}{#1}}

\newcommand{\addedok}[1]{\textcolor{cyan}{#1}}

\newcommand{\remove}[1]{\textcolor{green}{#1}}

\newcommand{\removehide}[1]{}



\newif\ifcomments
\commentstrue

\ifcomments

\newcommand{\commentsm}[1]{\textcolor{purple}{({\bf SM:} #1)}}
\newcommand{\commentsmhide}[1]{}

\newcommand{\commentec}[1]{\textcolor{magenta}{({\bf EC:} #1)}}
\newcommand{\commentdhhide}[1]{}

\newcommand{\commentps}[1]{\textcolor{orange}{({\bf PS:} #1)}}
\newcommand{\commentnwhide}[1]{}

\else

\newcommand{\commentsm}[1]{}
\newcommand{\commentsmhide}[1]{}

\newcommand{\commentec}[1]{}
\newcommand{\commentechide}[1]{}

\newcommand{\commentps}[1]{}
\newcommand{\commentpshide}[1]{}

\fi

%% file: appendix.tex

 




\section{Proofs}

\begin{proof}[Proposition~\ref{prop:completeness}]
We utilize the notion of consistent labelling and the completeness theorem (Theorem 1) proved in \cite{shati2021sat} in order to provide a constructive proof for Proposition~\ref{prop:completeness}.

Note that the precondition of the proposition is logically equivalent to the consistent labelling condition in \cite{shati2021sat}, when we rename $\Theta_C$ to $\gamma$. Given this precondition, Theorem 1 states that there exists a complete tree $\mathcal{D}$ of depth $d$ which gives the same labels as $\Theta_C$ to each point. Concluding our proof of the statement $\forall x\in X: \Theta_\mathcal{D}(x) = \Theta_C(x)$.
\end{proof}

\begin{proof}[Proposition~\ref{prop:eps_approx}]
We separately prove the first and second part of the Proposition~\ref{prop:eps_approx}.

\begin{enumerate}
    \item Let $\lambda^-_{min}$ be the lowest $\lambda^-$ value possible and $\mathcal{D}^*$ be the solution yielded by this minimization. In order to prove by contradiction, we assume that there exists another solution $\mathcal{D}$ which improves $MD$ by more than $\epsilon$.
    
    $$MD_\mathcal{D} < MD_{\mathcal{D}^*} - \epsilon$$
    
    First, we want to show that the pair $(x^{D^*}_1,x^{D^*}_2)$ corresponding to the distance $MD_\mathcal{D^*}$ belongs to $D_{\lambda^-_{min}}$.
    
    \begin{itemize}
        \item If $(x^{D^*}_1,x^{D^*}_2)\in D_w$ with $w > \lambda^-_{min}$, we have $\Theta_\mathcal{D^*}(x_1) \neq \Theta_\mathcal{D^*}(x_2)$ which contradicts the fact that $(x_1, x_2)$ is a diameter.
        \item If $(x^{D^*}_1,x^{D^*}_2)\in D_w$ with $w < \lambda^-_{min}$, we can conclude that all pairs belonging to $MD_\mathcal{D^*}$ are clustered separately since $(x^{D^*}_1,x^{D^*}_2)$ is the longest diameter. Thus, $\lambda^-_{min}-1$ is also a valid $\lambda^-$ value, contradicting our assumption of optimality.
    \end{itemize}
    
    Since $(x^{D^*}_1,x^{D^*}_2)\in D_{\lambda^-_{min}}$, $MD_\mathcal{D} < MD_\mathcal{D^*} - \epsilon$, and considering the fact that the shortest and longest distances of each distance class are at most $\epsilon$ apart, we can conclude that the pair $(x^{D}_1,x^{D}_2)$ corresponding to the distance $MD_\mathcal{D}$ is in a distance class $D_{\lambda^-_c}$ with $\lambda^-_c<\lambda^-_{min}$. Since $(x^{D}_1,x^{D}_2)$ is the longest diameter, we can conclude:
    
    $$(x_1,x_2) \in D_w, w > \lambda^-_c \rightarrow \Theta_{\mathcal{D}}(x_1) \neq \Theta_{\mathcal{D}}(x_2)$$
    
    Showing that $\lambda^-_c$ is a valid $\lambda^-$ value for the solution $\mathcal{D}$, contradicting $\lambda^-_{min}$ being the lowest $\lambda^-$ value across all solutions. Therefore, our assumption of the existence of $\mathcal{D}$ is false and $MD_\mathcal{D^*}$ is at most $\epsilon$ apart from the optimal solution with regards to $MD$.

    \item Let $\lambda^-_{min}$ and $\lambda^-_{max}$ be Pareto optimal values for $\lambda^-$ and $\lambda^+$ and $\mathcal{D}^*$ be the solution yielded by this optimization. We want to show that there exists a solution $\mathcal{D}^P$ that is Pareto optimal with regards to minimizing $MD$ and maximizing $MS$ such that:
    \begin{align}
        & MD_{\mathcal{D}^P} \geq MD_{\mathcal{D}^*} - \epsilon  \\
        & MS_{\mathcal{D}^P} \leq MS_{\mathcal{D}^*} + \epsilon
    \end{align}
    
    We use reasoning analogous to that presented in Part 1, to show that there does not exist a solution $\mathcal{D}^D$ such that:
    \begin{align}
        & MD_{\mathcal{D}^D} < MD_{\mathcal{D}^*} - \epsilon \\
        & MS_{\mathcal{D}^D} \geq MS_{\mathcal{D}^*}
    \end{align}
    
    The same reasoning is used to show that there does not exist a solution $\mathcal{D}^S$ such that:
    \begin{align}
        & MD_{\mathcal{D}^S} \leq MD_{\mathcal{D}^*} \\
        & MS_{\mathcal{D}^S} > MS_{\mathcal{D}^*} + \epsilon
    \end{align}
    
    Note that instead of contradicting $\lambda^-_{min}$ being the minimum as in Part 1, we can show that the existence of these two solutions contradicts $\lambda^-_{min}$ and $\lambda^-_{max}$ being a pair of Pareto optimal solutions.
    
    Let $\mathcal{D}^P$ be the optimal solution to the following optimization problem alongside all of the constraints of the original problem:
    \begin{align}
        min: \, & MD-MS \\
        & MD \leq MD_{\mathcal{D}^*} \\
        & MS \geq MS_{\mathcal{D}^*}
    \end{align}
    
    The fact that $\mathcal{D}^P$ optimizes $MD-MS$ shows that it is Pareto optimal since $MD$ ($MS$) cannot be further decreased (increased). The constraints on the solution do not hinder Pareto optimality as one of them is a lower bound for $MS$ and the other an upper bound for $MD$.
    
    Note that $MD_{\mathcal{D}^P} < MD_{\mathcal{D}^*} - \epsilon$ leads to the existence of $\mathcal{D}^D$ and $MS_{\mathcal{D}^P} > MS_{\mathcal{D}^*} + \epsilon$ leads to the existence of $\mathcal{D}^S$, which both will result in contradictions.
    
    Therefore, $\mathcal{D}^P$ is our Pareto optimal solution that is no more than $\epsilon$ apart from $\mathcal{D}^*$ in either direction.
    
\end{enumerate}

\end{proof}

\section{Mixed Integer Programming Encoding}

In the following, we provide the extended mixed-integer optimization (MIO) formulation for clustering trees with constraints, based on \citeauthor{bertsimas2021interpretable}'s \shortcite{bertsimas2021interpretable} MIO formulation for clustering trees.

\paragraph{Variables.}

\begin{description}
    \item [~~$s_{i}$:]
    The $s(i)$ variables in the Silhouette objective.
    \item [~~$q_{i}$:]  
    The $b(i)$ variables in the Silhouette objective, namely the average distance between points and their second closest cluster.
    \item [~~$r_{i}$:]  
    The $a(i)$ variables in the Silhouette objective, namely the average distance between points and their cluster.
    \item [~~$m_{i}$:] 
    $max(b(i),a(i))$ in the Silhouette objective.
    \item [~~$\gamma_{it}$:] (Binary) 
    Cluster (leaf) $t$ is the 
    2nd closest cluster to $x_i$.
    \item [~~$c_{it}$:] 
    The distance between $x_i$ and cluster (leaf) $t$.
    \item [~~$z_{it}$:] (Binary)
    $x_i$ belongs to cluster (leaf) $t$.
    \item [~~$K_{t}$:]  
    The size of cluster (leaf) $t$.
    \item [~~$a_{jt}$:] (Binary)
    Feature $j$ is chosen at branching node $t$.
    \item [~~$d_{t}$:] (Binary)
    Branching node $t$ is activated.
    \item [~~$b_{t}$:]
    Threshold at branching node $t$.
    \item [~~$l_{t}$:] (Binary)
    Leaf node $t$ is activated.
\end{description}

\paragraph{Constraints.} The following set of linear and quadratic constraints model the objective (Eq.\ \eqref{eq:mip_1}) and guarantee the soundness with regards to the relation between Silhouette values (Eqs.\ \eqref{eq:mip_2}-\eqref{eq:mip_4}), the selection and distance to the second closest cluster (Eqs.\ \eqref{eq:mip_5}-\eqref{eq:mip_7}), the distance to each cluster (Eqs.\ \eqref{eq:mip_8}-\eqref{eq:mip_9}), the size of each cluster (Eqs.\ \eqref{eq:mip_10}-\eqref{eq:mip_11}), the activation of branching nodes (Eqs.\ \eqref{eq:mip_12}-\eqref{eq:mip_14}), the activation of leaf nodes (Eqs.\ \eqref{eq:mip_15}-\eqref{eq:mip_16}), and the splits leading up to appearance at leaves (Eqs.\ \eqref{eq:mip_17}-\eqref{eq:mip_19}).

\begin{align}
& \text{minimize} \quad -\frac{1}{|X|}\sum_{x_i\in X} s_i & \label{eq:mip_1} \\ \hline
& \text{subject to} \quad s_i m_i=q_i-r_i, & \forall x_i\in X \label{eq:mip_2} \\ \hline
& m_i \geq q_i, & \forall x_i\in X \label{eq:mip_3} \\ \hline
& m_i \geq r_i, & \forall x_i\in X \label{eq:mip_4} \\ \hline
& q_i = \sum_{t\in \mathcal{T}_L} \gamma_{it}c_{it}, & \forall x_i\in X \label{eq:mip_5} \\ \hline
& \sum_{t\in\mathcal{T}_L} \gamma_{it} = 1, & \forall x_i\in X \label{eq:mip_6} \\ \hline
& \gamma_{it} \leq (1-z_{it}), & \forall x_i\in X, t\in\mathcal{T}_L \label{eq:mip_7} \\ \hline
& r_i = \sum_{t\in \mathcal{T}_L} c_{it} z_{it}, & \forall x_i\in X, t\in \mathcal{T}_L \label{eq:mip_8} \\ \hline
& c_{it} K_t = \sum_{x_j\in X} z_{jt}|x_i - x_j|, & \forall x_i\in X, t \in \mathcal{T}_L \label{eq:mip_9} \\ \hline
& \gamma_{it} \leq K_t, &  \forall x_i \in X, t \in \mathcal{T}_L \label{eq:mip_10} \\ \hline
& K_t = \sum_{x_i\in X} z_{it} & \forall t\in \mathcal{T}_L \label{eq:mip_11} \\ \hline
& \sum_{j \in F} a_{jt} = d_t, & \forall t\in \mathcal{T}_B \label{eq:mip_12} \\ \hline
& 0 \leq b_t \leq 100 d_t, & \forall t\in \mathcal{T}_B \label{eq:mip_13} \\ \hline
& d_t \leq d_{p(t)}, & \forall t\in \mathcal{T}_B -\{1\} \label{eq:mip_14} \\ \hline
& z_{it} \leq l_t, & \forall t\in \mathcal{T}_L \label{eq:mip_15} \\ \hline
& \sum_{x_i\in X} z_{it} \geq N_{min}l_t, & \forall t\in \mathcal{T}_L \label{eq:mip_16} \\ \hline
& \sum_{t\in \mathcal{T}_L} z_{it} = 1, & \forall x_i\in X \label{eq:mip_17} \\ \hline
\begin{split}
\mathrlap{a_m^T x_i \geq b_m - 100(1-z_{it}),} \\
\mathrlap{\forall x_i\in X, t\in \mathcal{T}_L, m \in A_R(t)} \label{eq:mip_18}
\end{split} \\ \hline
\begin{split}
\mathrlap{a_m^T x_i + \epsilon_s \leq  b_m + (100+\epsilon_{s})(1-z_{it}),} \\
\mathrlap{\forall x_i\in X, t\in\mathcal{T}_L, m \in A_L(t)} \label{eq:mip_19}
\end{split} \\ \hline
& z_{it} = z_{jt}, & \forall t\in \mathcal{T}_L, (i,j)\in ML \label{eq:mip_24} \\ \hline
& \sum_{t\in \mathcal{T}_L} (z_{it} - z_{jt})(z_{it} - z_{jt}) \geq 1, & \forall (i,j)\in CL \label{eq:mip_25} \\ \hline
& a_{jt}, d_t \in \{0,1\}, & \forall j\in F, t\in \mathcal{T}_B \label{eq:mip_20} \\ \hline
& z_{it}, l_t \in \{0,1\}, & \forall x_i \in X, t\in \mathcal{T}_L \label{eq:mip_21} \\ \hline
& \gamma_{it} \in \{0,1\}, & \forall x_i\in X, t\in \mathcal{T}_L \label{eq:mip_22} \\ \hline
& b_t \leq 100, & \forall t\in \mathcal{T}_B \label{eq:mip_23}
\end{align}

where $N_{min}$ is the minimum support for the number of points appearing at each active leaf. See \citeauthor{bertsimas2021interpretable} \shortcite{bertsimas2021interpretable} for a detailed discussion on the role of each constraint.


\paragraph{Changes from Bertsimas \textit{et al.} [2021].} We made a series of changes to the original MIO model as presented in \citeauthor{bertsimas2021interpretable} \shortcite{bertsimas2021interpretable}. We have categorized the modifications into three groups based on their nature and significance.

\begin{itemize}
    \item Changes in the notation:
    \begin{itemize}
        \item Iterating over points: $i=1,...,n \rightarrow x_i \in X$
        \item Iterating over features: $j=1,...,p \rightarrow f \in F$
        \item Distance between two points: $d_{ij} \rightarrow |x_i - x_j|$
        \item The epsilon value used in modelling splits (to avoid confusion with the epsilon used for approximating objective in our method): $\epsilon \rightarrow \epsilon_s$
    \end{itemize}
    \item Fixing typographical errors:
    \begin{itemize}
        \item The first dimension of the $z_{it}$ in Eq.\ \eqref{eq:mip_21}: $p \rightarrow n$
        \item Sum index in Eq.\ \eqref{eq:mip_8}: $\forall t \in T_L \rightarrow t \in T_L$
        \item Misuse of branching nodes in Eqs.\ \eqref{eq:mip_18} and \eqref{eq:mip_19}: $T_B \rightarrow T_L$
        \item Index of $b$ variables in Eqs.\ \eqref{eq:mip_18} and \eqref{eq:mip_19}: $b_t \rightarrow b_m$
    \end{itemize}
    \item Changes in the model:
    \begin{itemize}
        \item The original model does not support must-links and cannot-links: added Eqs.\ \eqref{eq:mip_24} and \eqref{eq:mip_25}.
        \item The big M constant in Eq.\ \eqref{eq:mip_7} serves no purpose: removed $M$.
        \item Gurobi does not allow division to be used in Eqs.\ \eqref{eq:mip_2} and \eqref{eq:mip_9}: replaced divisions with their equivalent multiplication. Note that the divided-by-zero cases should be monitored to maintain the correct semantics.
        \item Eq.\ \eqref{eq:mip_5} allowing $q_i$ to be arbitrarily large will trivialize the objective: $\geq \rightarrow =$
        \item Not having Eq.\ \eqref{eq:mip_10} will allow empty clusters to be selected as the second closest cluster: added Eq.\ \eqref{eq:mip_10}
        \item If $\epsilon_s$ is multiplied by $a_m^T$ on the left hand side of Eq.\ \eqref{eq:mip_19}, it would be possible for points to be directed to both left and right children, if a branching node is disabled: $a_m^T (x_i + \epsilon_s) \rightarrow a_m^T x_i + \epsilon_s$
        \item Constraints Eqs.\ \eqref{eq:mip_13}, \eqref{eq:mip_18}, and \eqref{eq:mip_19} need to be modified for a feature normalization to $[0,100]$ rather than $[0,1]$: $d_t \rightarrow 100 d_t$, $(1-z_{it}) \rightarrow 100(1-z_{it})$, and $(1+\epsilon_{s}) \rightarrow (100+\epsilon_{s})$
        \item Not having a constant upper bound for $b_t$ variables caused runtime issues: added Eq.\ \eqref{eq:mip_23}
    \end{itemize}
\end{itemize}

\section{Smart Pairs Algorithm}

In Algorithm~\ref{alg:cap}, we provide a detailed description of how inner and crossing edges (definition~\ref{def:inner_cross_edge}) can be utilized in the smart pairs procedure for detecting redundant clauses and potential infeasibility. 

\begin{algorithm}[t!]
\caption{Smart Pairs}\label{alg:cap}
\small
\begin{algorithmic}[1]
\State $E^+ = E^- = \emptyset$
\For{$(x_{i},x_{i'})\in ML$ sorted in ascending $<^*$}
    \If{$(x_{i},x_{i'})$ is not \textit{inner}} 
        \State $E^+ = E^+ \cup \{(x_{i},x_{i'})\}$
        \State add clauses Eq. (\ref{eq:19},~\ref{eq:20}) for $(x_{i},x_{i'})$
    \EndIf
\EndFor

\For{$(x_{i},x_{i'})\in CL$ sorted in descending $<^*$}
    \If{$(x_{i},x_{i'})$ is \textit{inner}}
        \State \text{Return Infeasible}
    \EndIf
    \If{$(x_{i},x_{i'})$ is not \textit{crossing}} 
        \State $E^- = E^- \cup \{(x_{i},x_{i'})\}$
        \State add clauses Eq. (\ref{eq:16},~\ref{eq:17},~\ref{eq:18}) for $(x_{i},x_{i'})$
    \EndIf
\EndFor

\State $\hat{E}^+ = E^+$

    \For{$(x_{i},x_{i'}) \in X\times X$ sorted in ascending $<^*$}
        \State set $w$ s.t. $(x_{i},x_{i'})\in D_w$
        \If{$(x_{i},x_{i'})$ is \textit{crossing}}
            \State add clause $(\lnot b^+_w)$
            \State Break
        \EndIf
        
        \If{$(x_{i},x_{i'})$ is not \textit{inner}}
            \State $E^+ = E^+ \cup \{(x_{i},x_{i'})\}$
            \State add clauses Eq. (\ref{eq:24},~\ref{eq:25}) for $(x_{i},x_{i'})$ and $w$
        \EndIf
    \EndFor

\State $E^+ = \hat{E}^+$ \Comment{Discard implied pairs for cond. co-clusering}

    \For{$(x_{i},x_{i'}) \in X\times X$ sorted in descending $<^*$}
        \State set $w$ s.t. $(x_{i},x_{i'})\in D_w$    
        \If{$(x_{i},x_{i'})$ is \textit{inner}}
            \State add clause $(b^-_w)$
            \State Break
        \EndIf
        
        \If{$(x_{i},x_{i'})$ is not \textit{crossing}}
            \State $E^- = E^- \cup \{(x_{i},x_{i'})\}$
            \State add clauses Eq. (\ref{eq:21},~\ref{eq:22},~\ref{eq:23}) for $(x_{i},x_{i'})$ and $w$
        \EndIf
    \EndFor
\end{algorithmic}
\end{algorithm}

\section{NMI Results}

\begin{table*}[htp!]
    \centering
    \caption{NMI scores for interpretable constrained clustering with $\epsilon=0.1$ averaged over $20$ runs.}
    \label{tab:tree_nmi}
    \small
    \begin{tabular}{cc|cccc|cccc}
    \toprule
\multirow{2}{*}{Dataset} & \multirow{2}{*}{$\kappa$} & \multicolumn{4}{c}{[MD,MS]} & \multicolumn{4}{c}{MD}\\
 & & NMI & NMI (F) & Feas. & Time (s) & NMI & NMI (F) & Feas. & Time (s) \\
 \midrule
Iris & 0.0 & 0.67 & 0.67 & 20 & 0.6 & 0.68 & 0.72 & 20 & 0.4 \\
$\vert X \vert=150$ & 0.1 & 0.83 & 0.79 & 20 & 0.7 & 0.72 & 0.6 & 20 & 0.4 \\
$\vert F \vert=4$ & 0.25 & 0.85 & 0.79 & 20 & 0.7 & 0.8 & 0.62 & 20 & 0.4 \\
$\vert C \vert=3$ & 0.5 & 0.89 & 0.79 & 20 & 0.8 & 0.86 & 0.65 & 20 & 0.5 \\
$d=3$ & 1.0 & 0.93 & 0.89 & 16 & 0.7 & 0.93 & 0.71 & 16 & 0.4 \\
\hline Wine & 0.0 & 0.02 & 0.02 & 20 & 0.7 & 0.41 & 0.2 & 20 & 0.6 \\
$\vert X \vert=178$ & 0.1 & 0.68 & 0.53 & 20 & 1.2 & 0.44 & 0.22 & 20 & 0.6 \\
$\vert F \vert=13$ & 0.25 & 0.76 & 0.57 & 20 & 1.8 & 0.59 & 0.23 & 20 & 0.9 \\
$\vert C \vert=3$ & 0.5 & 0.79 & 0.49 & 20 & 6.9 & 0.69 & 0.22 & 20 & 5.4 \\
$d=3$ & 1.0 & 0.9 & 0.69 & 20 & 8.8 & 0.86 & 0.25 & 20 & 8 \\
\hline Glass & 0.0 & 0.37 & 0.38 & 20 & 2.1 & 0.32 & 0.38 & 20 & 1.2 \\
$\vert X \vert=214$ & 0.1 & 0.3 & 0.2 & 20 & 9.3 & 0.27 & 0.21 & 20 & 3.3 \\
$\vert F \vert=9$ & 0.25 & 0.32 & 0.18 & 20 & 61.1 & 0.25 & 0.16 & 20 & 35.1 \\
$\vert C \vert=7$ & 0.5 & 0.32 & 0.18 & 19 & 954.1 & 0.32 & 0.13 & 20 & 624.9 \\
$d=4$ & 1.0 & - & 0.2 & 0 & 193.8 & - & 0.14 & 0 & 200.1 \\
\hline Ionosphere & 0.0 & 0.02 & 0.02 & 20 & 2.4 & 0.09 & 0.07 & 20 & 2 \\
$\vert X \vert=351$ & 0.1 & 0.21 & 0.11 & 20 & 33.7 & 0.09 & 0.07 & 20 & 10.2 \\
$\vert F \vert=34$ & 0.25 & 0.39 & 0.2 & 11 & 598.9 & 0.37 & 0.08 & 11 & 880.9 \\
$\vert C \vert=2$ & 0.5 & - & 0.32 & 0 & 29.5 & - & 0.13 & 0 & 31.4 \\
$d=3$ & 1.0 & - & 0.7 & 0 & 5.8 & - & 0.67 & 0 & 5.7 \\
\hline Seeds & 0.0 & 0.65 & 0.63 & 20 & 1.8 & 0.57 & 0.68 & 20 & 0.5 \\
$\vert X \vert=210$ & 0.1 & 0.65 & 0.63 & 20 & 1 & 0.65 & 0.56 & 20 & 0.5 \\
$\vert F \vert=7$ & 0.25 & 0.7 & 0.63 & 20 & 1.3 & 0.69 & 0.55 & 20 & 0.7 \\
$\vert C \vert=3$ & 0.5 & 0.74 & 0.63 & 14 & 1.5 & 0.74 & 0.54 & 14 & 1.2 \\
$d=3$ & 1.0 & 0.87 & 0.77 & 1 & 0.7 & 0.9 & 0.7 & 1 & 0.7 \\
\hline Libras & 0.0 & 0.46 & 0.49 & 20 & 1802.8 & 0.47 & 0.39 & 20 & 1802.9 \\
$\vert X \vert=360$ & 0.1 & 0.42 & 0.42 & 20 & 1159.1 & 0.4 & 0.34 & 20 & 941.6 \\
$\vert F \vert=90$ & 0.25 & 0.43 & 0.38 & 20 & 1080 & 0.38 & 0.32 & 20 & 681.9 \\
$\vert C \vert=15$ & 0.5 & 0.43 & 0.33 & 20 & 866.2 & 0.37 & 0.26 & 20 & 305.1 \\
$d=5$ & 1.0 & 0.42 & 0.32 & 20 & 1518.3 & 0.37 & 0.24 & 20 & 859.4 \\
\hline Spam & 0.0 & 0 & 0 & 20 & 38 & 0.04 & 0 & 20 & 41.5 \\
$\vert X \vert=4601$ & 0.1 & - & 0.02 & 0 & 358.9 & - & 0.03 & 0 & 379.4 \\
$\vert F \vert=57$ & 0.25 & - & 0.05 & 0 & 286.1 & - & 0.1 & 0 & 312.6 \\
$\vert C \vert=2$ & 0.5 & - & 0.06 & 0 & 151.1 & - & 0.2 & 0 & 152.9 \\
$d=3$ & 1.0 & - & 0.45 & 0 & 77.1 & - & 0.45 & 0 & 77.9 \\
\hline Lsun & 0.0 & 0.54 & 0.49 & 20 & 1.2 & 0.48 & 0.5 & 20 & 0.8 \\
$\vert X \vert=400$ & 0.1 & 0.95 & 0.65 & 20 & 1 & 0.73 & 0.28 & 20 & 0.7 \\
$\vert F \vert=2$ & 0.25 & 1 & 0.94 & 20 & 1 & 0.86 & 0.25 & 20 & 0.6 \\
$\vert C \vert=3$ & 0.5 & 1 & 1 & 20 & 0.9 & 0.93 & 0.25 & 20 & 0.6 \\
$d=3$ & 1.0 & 1 & 1 & 20 & 0.9 & 0.97 & 0.39 & 20 & 0.6 \\
\hline Chainlink & 0 & 0.09 & 1 & 20 & 3 & 0.1 & 0.07 & 20 & 2 \\
$\vert X \vert=1000$ & 0.1 & 0.81 & 1 & 17 & 5 & 0.76 & 0.01 & 17 & 5.5 \\
$\vert F \vert=3$ & 0.25 & 0.82 & 1 & 1 & 1.8 & 0.85 & 0.05 & 1 & 1.7 \\
$\vert C \vert=2$ & 0.5 & - & 1 & 0 & 1.3 & - & 0.05 & 0 & 1.2 \\
$d=3$ & 1.0 & - & 1 & 0 & 1 & - & 0.49 & 0 & 0.9 \\
\hline Target & 0.0 & 0.44 & 0.44 & 20 & 5.7 & 0.4 & 0.39 & 20 & 3.7 \\
$\vert X \vert=770$ & 0.1 & 1 & 1 & 20 & 2.8 & 0.6 & 0.55 & 20 & 18.8 \\
$\vert F \vert=2$ & 0.25 & 1 & 1 & 20 & 2.7 & 0.8 & 0.41 & 20 & 9.9 \\
$\vert C \vert=6$ & 0.5 & 1 & 1 & 20 & 2.4 & 0.91 & 0.26 & 20 & 4.4 \\
$d=4$ & 1.0 & 1 & 1 & 20 & 2.2 & 0.97 & 0.4 & 20 & 1.9 \\
\hline WingNut & 0.0 & 1 & 1 & 20 & 2.5 & 0.99 & 0.27 & 20 & 2 \\
$\vert X \vert=1016$ & 0.1 & 1 & 1 & 20 & 1.7 & 0.98 & 0.26 & 20 & 1.3 \\
$\vert F \vert=2$ & 0.25 & 1 & 1 & 20 & 1.6 & 0.99 & 0.31 & 20 & 1.2 \\
$\vert C \vert=2$ & 0.5 & 1 & 1 & 20 & 1.6 & 0.99 & 0.46 & 20 & 1.2 \\
$d=3$ & 1.0 & 1 & 1 & 20 & 1.6 & 1 & 0.75 & 20 & 1.2 \\
\bottomrule
\end{tabular}
\end{table*}

The results presented in Table~\ref{tab:tree_nmi} and Table~\ref{tab:infeas_nmi} mirror the ones in Table~\ref{tab:tree} and Table~\ref{tab:infeas} with NMI scores instead of ARI. The improvement trends we see in NMI scores resemble those of ARI. Namely, Pareto optimal objective achieves higher average NMI compared to the diameter only case, and tree clustering outperforms CC on average as well.

\begin{table}[t!]
    \centering
    \caption{NMI scores for tree clustering with [MD,MS] objective for different tree depths ($\kappa=0.5$, $\epsilon=0.1$) averaged over $20$ runs.}
    \label{tab:infeas_nmi}
    \small
    \begin{tabular}{cc|ccc}
    \toprule
Dataset & $d$ & NMI & Feas. & Time (s) \\
 \midrule
 \multirow{3}{*}{Iris} & 2 & 0.81 & 2 & 0.3 \\
& 3 & 0.89 & 20 & 0.8 \\
& 4 & 0.86 & 20 & 0.8 \\
\hline  \multirow{3}{*}{Wine} & 2 & 0.84 & 2 & 0.4 \\
& 3 & 0.79 & 20 & 6.9 \\
& 4 & 0.69 & 20 & 6.2 \\
\hline  \multirow{3}{*}{Glass} & 3 & - & 0 & 2.6 \\
& 4 & 0.32 & 19 & 953.6 \\
& 5 & 0.28 & 20 & 35.8 \\
\hline  \multirow{3}{*}{Ionosphere} & 2 & - & 0 & 0.8 \\
& 3 & - & 0 & 29.5 \\
& 4 & 0.57 & 18 & 731.7 \\
\hline  \multirow{3}{*}{Seeds} & 2 & - & 0 & 0.3 \\
& 3 & 0.74 & 14 & 1.5 \\
& 4 & 0.71 & 20 & 2.1 \\
\hline  \multirow{3}{*}{Libras} & 4 & 0.4 & 2 & 1801.4 \\
& 5 & 0.43 & 20 & 866.2 \\
& 6 & 0.41 & 20 & 315.4 \\
\hline  \multirow{3}{*}{Spam} & 2 & - & 0 & 35 \\
& 3 & - & 0 & 151.1 \\
& 4 & - & 0 & 975.6 \\
\hline  \multirow{3}{*}{Lsun} & 2 & 1 & 20 & 0.9 \\
& 3 & 1 & 20 & 0.9 \\
& 4 & 1 & 20 & 1.1 \\
\hline \multirow{3}{*}{Chainlink} & 2 & - & 0 & 0.8 \\
 & 3 & - & 0 & 1.3 \\
 & 4 & 1 & 20 & 3.4 \\
\hline  \multirow{3}{*}{Target} & 3 & - & 0 & 1.8 \\
& 4 & 1 & 20 & 2.5 \\
& 5 & 1 & 20 & 4 \\
\hline  \multirow{3}{*}{WingNut} & 2 & 1 & 20 & 1.5 \\
& 3 & 1 & 20 & 1.7 \\
& 4 & 1 & 20 & 2.1 \\
\bottomrule
\end{tabular}
\end{table}